\documentclass[letterpaper]{article} % DO NOT CHANGE THIS
\usepackage{aaai25}  % DO NOT CHANGE THIS
\usepackage{times}  % DO NOT CHANGE THIS
\usepackage{helvet}  % DO NOT CHANGE THIS
\usepackage{courier}  % DO NOT CHANGE THIS
\usepackage[hyphens]{url}  % DO NOT CHANGE THIS
\usepackage{graphicx} % DO NOT CHANGE THIS
\usepackage{natbib}  % DO NOT CHANGE THIS AND DO NOT ADD ANY OPTIONS TO IT
\usepackage{caption} % DO NOT CHANGE THIS AND DO NOT ADD ANY OPTIONS TO IT
\usepackage{algorithm}
\usepackage{algorithmic}

\usepackage{newfloat}
\usepackage{listings}
\DeclareCaptionStyle{ruled}{labelfont=normalfont,labelsep=colon,strut=off} % DO NOT CHANGE THIS
\floatstyle{ruled}
\newfloat{listing}{tb}{lst}{}
\floatname{listing}{Listing}
\title{OmniSR: Shadow Removal Under Direct and Indirect Lighting}
\author{
    Jiamin Xu\textsuperscript{\rm 1}, 
    Zelong Li\textsuperscript{\rm 1},
    Yuxin Zheng\textsuperscript{\rm 1},
    Chenyu Huang\textsuperscript{\rm 1},
    Renshu Gu\textsuperscript{\rm 1}, 
    Weiwei Xu\textsuperscript{\rm 2},   
    Gang Xu\textsuperscript{\rm 1}\thanks{Corresponding author.}
}
\affiliations{
    \textsuperscript{\rm 1}Hangzhou Dianzi University\\
    \textsuperscript{\rm 2}Zhejiang University\\
    superxjm@yeah.net,
    \{jokerli,yuxin6,shamet,renshugu\}@hdu.edu.cn,
    xww@cad.zju.edu.cn,
    gxu@hdu.edu.cn
}

\usepackage{bibentry}
\usepackage{amsmath}
\usepackage[dvipsnames]{xcolor}
\usepackage{subcaption}
\usepackage{todonotes}
\usepackage{multirow}
\usepackage{colortbl}
\usepackage{xcolor}
\usepackage{array}
\definecolor{colorTrd}{rgb}{0.95,0.95,0.65}
\definecolor{colorSnd}{rgb}{1, 0.85, 0.7}
\definecolor{colorFst}{rgb}{1, 0.7, 0.7}

\usepackage{amsfonts}
\usepackage{float}
\usepackage{multicol}
\usepackage{booktabs}

\begin{document}

\maketitle
\begin{abstract}

Shadows can originate from occlusions in both direct and indirect illumination. Although most current shadow removal research focuses on shadows caused by direct illumination, shadows from indirect illumination are often just as pervasive, particularly in indoor scenes. A significant challenge in removing shadows from indirect illumination is obtaining shadow-free images to train the shadow removal network. To overcome this challenge, we propose a novel rendering pipeline for generating shadowed and shadow-free images under direct and indirect illumination, and create a comprehensive synthetic dataset that contains over 30,000 image pairs, covering various object types and lighting conditions. We also propose an innovative shadow removal network that explicitly integrates semantic and geometric priors through concatenation and attention mechanisms. The experiments show that our method outperforms state-of-the-art shadow removal techniques and can effectively generalize to indoor and outdoor scenes under various lighting conditions, enhancing the overall effectiveness and applicability of shadow removal methods. 

\end{abstract}

\begin{links}
    \link{Code}{https://blackjoke76.github.io/Projects/OmniSR/}
    % \link{Datasets}{https://blackjoke76.github.io/Projects/OmniSR/}
    % \link{Extended version}{https://aaai.org/example/extended-version}
\end{links}    
\begin{figure}[t]
\centering
\includegraphics[width=\linewidth]{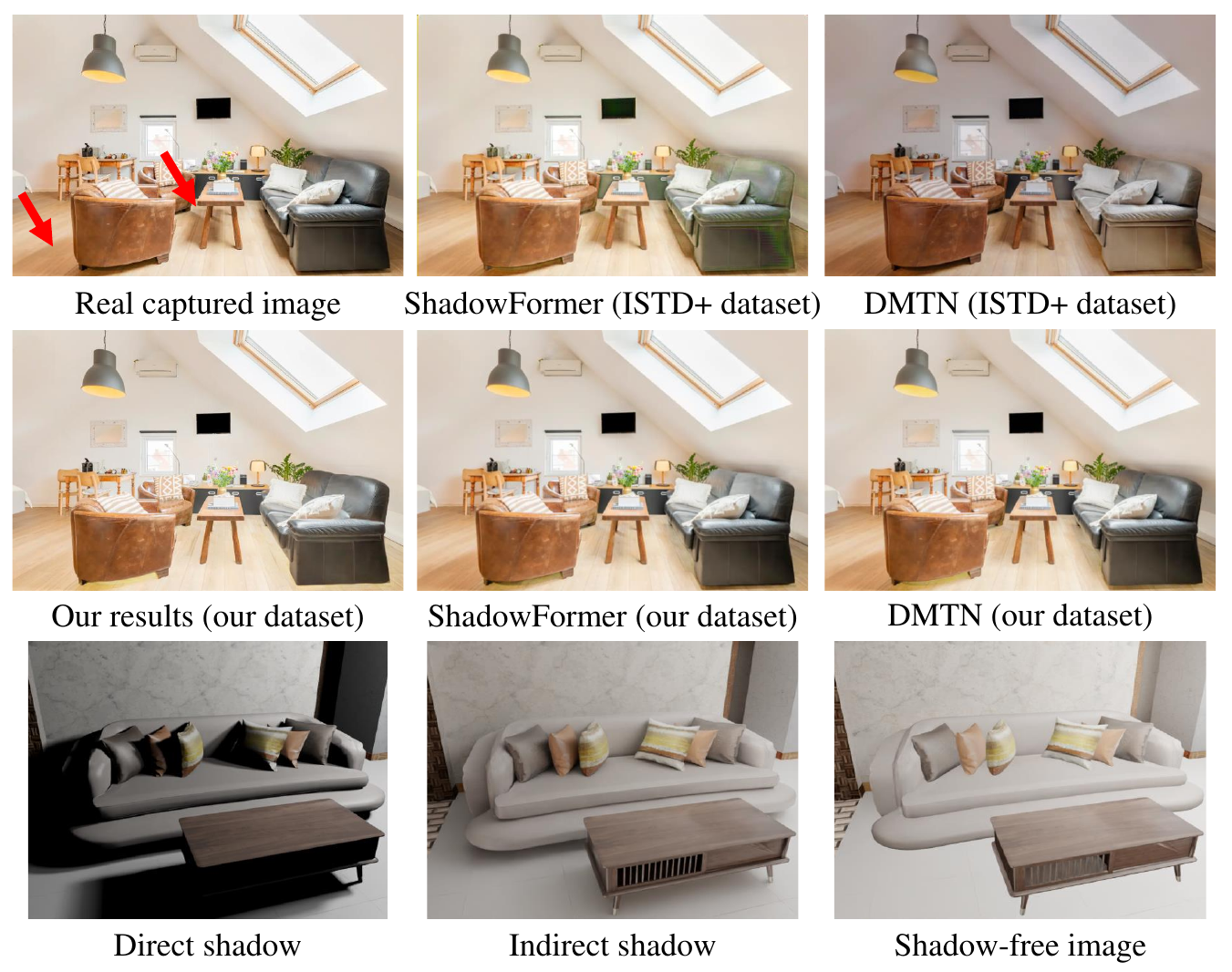}
\vspace{-0.5cm}
\caption{\emph{Top}: Removing shadows from intricate indoor scenes with indirect illumination poses a challenge for current methods such as ShadowFormer~\cite{guo2023shadowformer} and DMTN~\cite{liu2023decoupled}, especially when trained on existing datasets focusing on direct illumination, like ISTD+~\cite{le2019shadow}. In contrast, our shadow removal network, along with a newly introduced INdirect Shadow (\textbf{INS}) dataset, demonstrates a superior ability to remove shadows accurately. \emph{Bottom}: The rendering results reveal the prominence and prevalence of both direct and indirect shadows in indoor scenes.}
\vspace{-0.2cm}
\label{fig:teaser} 
\end{figure}

\section{Introduction}
\label{sec:intro}

Shadows are omnipresent phenomena that emerge due to occlusions within a scene's global illumination. Removing shadows is crucial, as it can enhance the performance of various computer vision tasks, such as object segmentation and tracking~\cite{he2017mask,sanin2010improved}, intrinsic decomposition~\cite{li2018learning,li2022physically,nestmeyer2020learning,ye2023intrinsicnerf}, and 3D reconstruction~\cite{ling2023shadowneus}. 

Recent shadow removal techniques predominantly adopt deep learning-based approaches, leveraging publicly available datasets containing pairs of shadowed and shadow-free images~\cite{wang2018stacked,le2019shadow,qu2017deshadownet} or shadow masks~\cite{vicente2016large,sun2023adaptive}. These datasets, derived from real-world captures, often suffer from limitations in both quantity and quality. To overcome these limitations, many researchers  employ techniques such as data augmentation~\cite{le2018a+,cun2020towards}, self-training~\cite{yang2023silt}, and unsupervised learning~\cite{jiang2023learning,hu2019mask,jin2021dc}. On the other hand, due to the visual similarity between certain shadowed and non-shadowed regions, many studies focus on enhancing the discriminative power of shadow removal networks. This can be achieved by incorporating semantic information~\cite{zheng2019distraction,hu2019direction,liu2023shadow}, employing binary shadow masks~\cite{li2023leveraging,guo2023shadowformer}, or adopting progressive shadow removal methods~\cite{zhu2022efficient,guo2023shadowdiffusion,ding2019argan}. Despite these advancements, existing methods still face challenges in removing shadows in complex indoor scenes (Fig.~\ref{fig:teaser}).

Indoor scene shadow removal is particularly challenging due to the complex nature of global illumination, which encompasses direct lighting\footnote{Light rays bounce once off an object's surface to reach the camera.} and indirect lighting. The indirect lighting, involving rays bouncing off surfaces multiple times before being captured, leads to a substantial proportion of shadows that appear soft and subtle, as depicted in Fig.~\ref{fig:teaser}. We distinguish shadows originating from direct lighting as ``Direct Shadows'' and those produced by indirect lighting as ``Indirect Shadows''. Despite the prevalence of indirect shadows in indoor scenes, existing research and datasets have largely neglected them. This bias can be attributed to three primary reasons: 1. It is difficult to define indirect shadows precisely; 2. It is not feasible to remove all occluders in complex indoor scenes, making it impossible to capture images completely free of both direct and indirect shadows; 3. Indirect shadows can be overlooked in most outdoor scenes. Unlike existing methods, our approach considers both direct and indirect shadows. To address the challenges above, we define and render shadow and shadow-free images within the framework of path tracing~\cite{Tomas2018Real-TimeRendering}. While shadow images are naturally generated in the path tracing pipeline, producing shadow-free images requires the separate definition of each shadow type and the design of distinct rendering procedures.

% Unlike existing methods, our approach considers both direct and indirect shadows. To address the challenges above, we define and render shadow and shadow-free images within the framework of path tracing~\cite{Tomas2018Real-TimeRendering}. In the path-tracing pipeline, shadow images are naturally generated. However, to produce shadow-free images, it is essential to separately define each type of shadow and design distinct rendering scripts to remove them.

% and create a high-quality synthetic dataset to enhance shadow removal tasks in indoor and outdoor environments.

To effectively remove both direct and indirect shadows from a single image, we propose a shadow removal network that integrates semantic and geometric information. Specifically, inspired by screen-space ambient occlusion~\cite{miller1994efficient}, we use depth information to guide the removal of indirect shadows, which commonly occur near depth discontinuities. Meanwhile, the incorporation of semantic information helps identify direct shadows, which typically appear darker than the surrounding areas with the same semantic content. Consequently, our network takes RGB-Depth (RGBD) as input, concatenates semantic features to the bottleneck layer, and reweights features in local attention blocks based on local semantic and geometric similarities. Unlike the shadow-interaction attention employed in ShadowFormer~\cite{guo2023shadowformer}, which exploits contextual correlations between binary shadowed and non-shadow regions, our network incorporates geometric information and semantic features and does not rely on shadow detection as auxiliary information. 

% Our critical insight is that shadows cast on a surface result in darker shading compared to the surrounding non-shadowed area with the same semantic properties. 

% Consequently, shadow detection implies an intensity comparison within the spatial context with similar semantics. Shadow removal can be achieved by inpainting the content of shadowed areas using surrounding areas with brighter intensity, similar semantics, and short geometric distances. 

% The semantic and geometric information provides more precise guidance for restoring a shadow-free image. 

In summary, the main contributions of this work are three-fold:
\renewcommand{\labelitemi}{$\bullet$}
\begin{itemize}%[leftmargin=12pt,topsep=4pt]
\item We propose a novel rendering pipeline for shadows that considers direct and indirect illumination and construct the first INdirect Shadow (\textbf{INS}) dataset consisting of 30,000+ synthetic image pairs with diverse scene types. 

% This dataset was generated using Cycles~\cite{blender}, Blender's ray-tracing-based production rendering engine, and leveraged the synthetic indoor scenes offered by 3D-FRONT~\cite{fu20213dfront,fu20213dfuture} as well as object models from the ABO~\cite{collins2022abo} and Objaverse~\cite{deitke2023objaverse} datasets.

\item We introduce a novel shadow removal network that efficiently removes both direct and indirect shadows. The network leverages semantic and geometric priors through RGBD input, incorporating semantic features, and semantic and geometry-aware attention mechanisms. 

% These elements are integrated into the Swin U-Net framework~\cite{liu2021swin,wang2022uformer}.

\item The extensive experimental results on our \textbf{INS} dataset and several other datasets demonstrate that our method has achieved a new state-of-the-art performance, particularly in handling shadows under both direct and indirect illumination. 

% We also assess the performance of our model using real indoor scene images containing indirect shadows. The results reveal that our method exhibits strong generalizability in real-world scenarios.
\end{itemize}

\section{Related Work}
\label{sec:related}

Single-image shadow removal aims to restore the shadow-free texture in shadowed regions while maintaining the appearance of non-shadowed areas. Traditional model-based approaches depend on the physical models of shadow images, but their reliance on prior knowledge often limits their effectiveness in real-world scenes~\cite{finlayson2009entropy,khan2015automatic,zhang2015shadow}. 

% , where dark non-shadow regions and relatively bright shadow regions often coexist.
In recent years, deep learning-based methods have shown remarkable performance in shadow removal thanks to their end-to-end capabilities. However, a key challenge for these methods lies in ensuring generalizability given the limitations in data quantity and quality, as well as resolving shadow ambiguity in real scenes. To address these challenges, researchers have explored various strategies, including the integration of multi-context features~\cite{qu2017deshadownet,hu2019direction,chen2021canet} and the use of generative adversarial networks~\cite{wang2018stacked,cun2020towards,hu2019mask}, to enhance generalizability and improve the effectiveness of available datasets. Other works, such as Fu et al.~\shortcite{fu2021auto} and Zhu et al.~\shortcite{zhu2022bijective}, treated shadow removal as an exposure fusion problem or developed a bijective mapping that combines the learning processes of shadow removal and generation.

More recently, Guo et al.~\shortcite{guo2023shadowformer} introduced ShadowFormer, a single-stage shadow removal network that utilizes multi-scale attention to exploit contextual correlations between shadowed and non-shadowed regions. Although lightweight, this network relies on the shadow mask as input, making its performance heavily dependent on high-precision shadow detection results. DMTN~\cite{liu2023decoupled}, introduced a decoupled multi-task network that learns decomposed features for shadow removal. Other methods, such as ShadowDiffusion~\cite{guo2023shadowdiffusion} and DeS3~\cite{jin2024des3} utilized diffusion models to remove shadows by treating the shadowed image as a condition. While diffusion-based methods have demonstrated efficacy in producing realistic shadow-free results, they suffer from high variance and are time-consuming during diffusion inference. ShadowRefiner~\cite{dong2024shadowrefiner} introduce a novel mask-free model that integrates spatial and frequency domain representations for image shadow removal, which achieves the champion in the Perceptual Track on NTIRE 2024 Image Shadow Removal Challenge~\cite{vasluianu2024ntire}.
% and achieves the second best performance in the Fidelity Track 

Existing deep learning-based methods face two main challenges. Firstly, they are primarily trained on datasets with little or no indirect illumination. As a result, their performance significantly degrades when dealing with complex and subtle shadows caused by indirect illumination. Secondly, few methods incorporate geometric information into shadow removal tasks. This may because most current datasets position shadow-casting objects outside the field of view to capture shadow-free images. However, we believe that geometric information is crucial for identifying shadows, as indirect shadows are often associated with depth discontinuities. In this work, we present a novel dataset for shadow removal under complex illumination. Additionally, we introduce a semantic and geometric-aware network that can effectively removes both direct and indirect shadows.

\section{Proposed Method}

% 和Secondary Light-source 的Distribution 有关
% In this section, we first define direct and indirect shadows based on occlusion and light intensity. According to the rendering equation, the final rendered image can be viewed as the sum of the direct illumination image and indirect illumination image \footnote{The final image needs to undergo non-linear gamma correction.}, represented as $\mathbf{I}_s = \mathbf{I}^{dr}_s + \mathbf{I}^{idr}_s$. Similarly, the shadow-free image is the combination of the direct and indirect shadow-free images, denoted as $\mathbf{I}_f = \mathbf{I}^{dr}_f + \mathbf{I}^{idr}_f$.

\subsection{Direct and Indirect Shadows}

Following the rendering equation~\cite{Tomas2018Real-TimeRendering}, each rendered image $\mathbf{I}_s$ can be viewed as the sum of an image with only direct illumination and an image with only indirect illumination. As both of these images inherently contain shadows, we refer to them as direct shadow image $\mathbf{I}^{dr}_s$ and indirect shadow image $\mathbf{I}^{idr}_s$, respectively. As is widely known, the direct shadows in $\mathbf{I}^{dr}_s$ arise from occlusions between shading points and light sources. Therefore, for each direct shadow image $\mathbf{I}^{dr}_s$, we can generate its corresponding image free of direct shadows, labelled as $\mathbf{I}^{dr}_f$, by removing any occlusions along the path of direct lighting.

\begin{figure}[t]
\centering
\includegraphics[width=\linewidth]{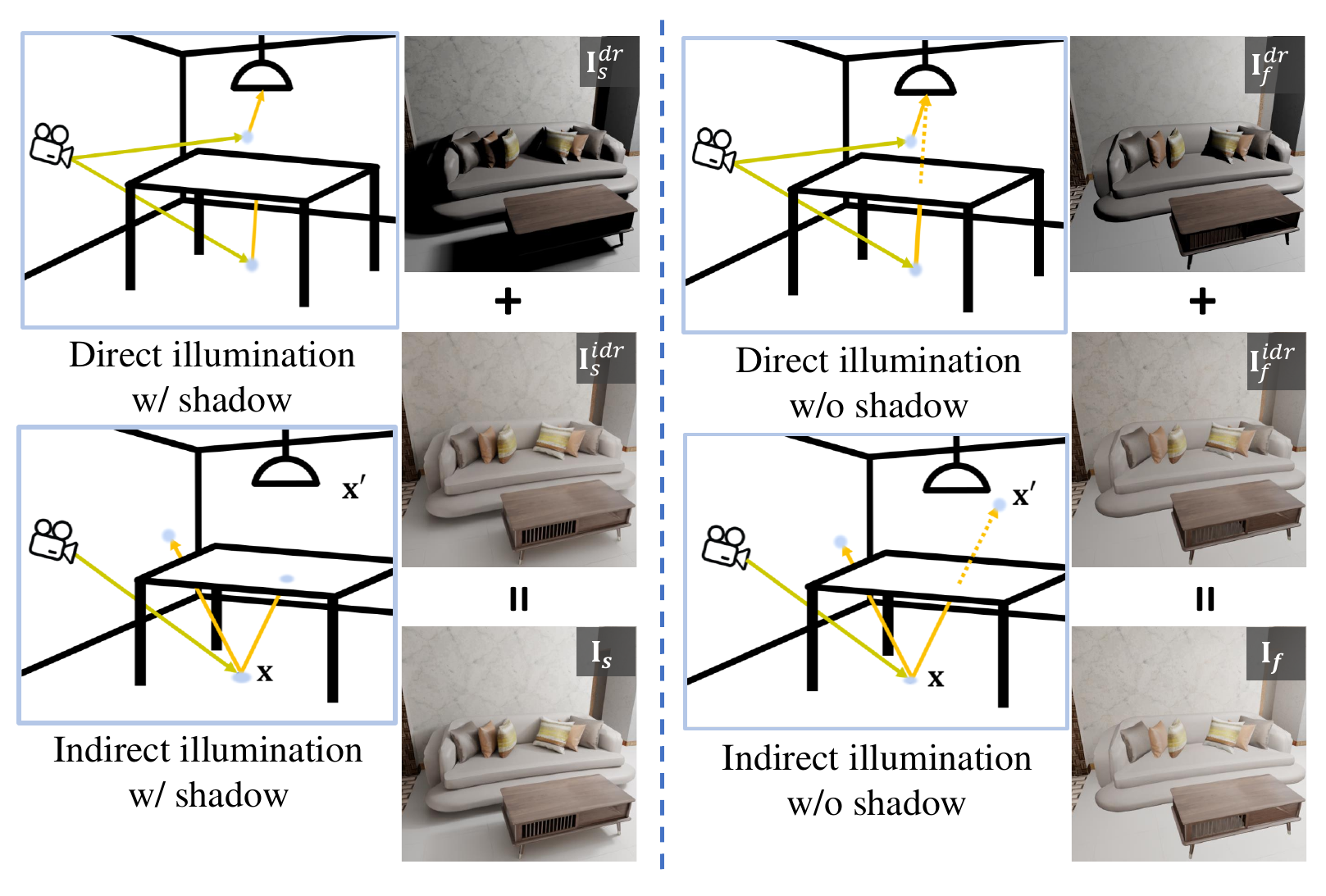}
\caption{\textbf{Rendering of shadow and shadow-free images.} \emph{Left}: The rendered image with shadow is a combination of a direct illumination image and an indirect illumination image. \emph{Right}: The final shadow-free image is a composite of a direct shadow-free image, and an indirect shadow-free image.}
\vspace{-0.5cm}
\label{fig:shadow_rendering} 
\end{figure}

On the contrary, precisely defining indirect shadows is challenging because, in $\mathbf{I}^{idr}_s$, the intensity of each shading point is influenced by ``secondary light sources'' from all directions. In some cases, the nearby objects can occlude the distant ``secondary light sources'', resulting in indirect shadows. For instance, in Fig.~\ref{fig:shadow_rendering}, the light incident on the point $\mathbf{x}$ from the background walls is blocked by the table, casting a shadow on the ground beneath the table. However, the bottom of the table also serves as a ``secondary light source''  capable of illuminating $\mathbf{x}$. In cases where the bottom of the table is brighter than the wall, e.g., there are light sources near the floor, we do not consider shading point $\mathbf{x}$ to contain shadows. As a result, we define indirect shadows as \emph{a darker shading result caused by occlusions from nearby objects compared to scenarios without such occlusion}. 

% This definition also aligns with Ambient Occlusion~(AO)~\cite{miller1994efficient}, which can be viewed as an approximation of indirect shadows under distant light sources.

This definition of indirect shadows is based on the principles of global illumination. In global illumination, Ambient Occlusion (AO)~\cite{miller1994efficient} is commonly used to approximate indirect shadows. AO ignores the varying intensity of indirect illumination in each direction and estimates shadows by counting nearby occlusions. We extend the concept of AO by incorporating the intensity of each secondary light sources and treating nearby occlusions as transparent, thereby achieving images free of indirect shadows.

With the aforementioned definition, we utilize a modified path tracing pipeline to render $\mathbf{I}^{dr}_s$, $\mathbf{I}^{idr}_s$, $\mathbf{I}^{dr}_f$, and $\mathbf{I}^{idr}_f$ respectively. Specifically, we generate $\mathbf{I}^{dr}_s$ by considering only the light directly originating from a light source or undergoing a single bounce (reflection or refraction) from a light source, as depicted in Fig. \ref{fig:shadow_rendering}. For $\mathbf{I}^{idr}_s$, we take into account light paths excluding the single bounce direct lighting from light sources. To generate the direct shadow-free image $\mathbf{I}^{dr}_f$, our approach eliminates occlusions along the first bounce's ray in $\mathbf{I}^{dr}_s$ to remove all direct shadows. Regarding the indirect shadow-free image, we first render $\mathbf{I}^{idr}_f$ by eliminating occlusions along the first bounce's ray within a radius of $r$. We empirically found that $r=1$ meter works well for our scenes (see supp. for rendering results with different $r$). We then compare the intensity of each pixel in $\mathbf{I}^{idr}_f$ with its corresponding pixel in $\mathbf{I}^{idr}_s$ and select the brighter intensity as the final indirect shadow-free image $\mathbf{I}^{idr}_f$. Note that the first intersection at the first bounce from a point will be preserved if it is the last intersected object, even when its distance is within 1 meter, such as walls or ceilings.

By incorporating indirect illumination and employing a novel approach to render indirect shadow-free images, our overall shadow-free image rendering demonstrates enhanced realism compared to the sole rendering of direct shadow-free images, as illustrated in Fig.~\ref{fig:shadow_rendering}. 

% After obtaining shadow image $\mathbf{I}_s=\mathbf{I}^{dr}_s+\mathbf{I}^{idr}_s$ and shadow-free image $\mathbf{I}_f=\mathbf{I}^{dr}_f+\mathbf{I}^{idr}_f$, the probabilistic shadow mask $\mathbf{S}$ is defined as:
% \begin{align}
% \label{equ:shadow_prob}
% \mathbf{S}=\mathrm{clip}(\left( \mathbf{I}_f-\mathbf{I}_s \right) /\mathrm{max}(\mathbf{I}_f,0.1)).
% \end{align}
% Here, $\mathrm{clip}(\cdot)$ clips the results within the range of $[0, 1]$. For real images with noise, a clipping threshold of 0.1 is applied to $\mathbf{I}_f$ to prevent the near-black areas from being easily misclassified as shadows.

\subsection{Our INS Dataset}

Based on the definition provided above, we implemented the corresponding direct/indirect shadow and shadow-free rendering pipeline using Blender Cycles engine \cite{blender}, with the help of Open Shading Language (OSL). The resulting collection of shadow and shadow-free images is referred to as the ``INS dataset''. The dataset includes 30,000 training and 2,000 testing images, all with a resolution of $512\times512$. The training and testing images are generated from distinct scenes with different objects and materials.

\begin{figure}[t]
\centering
\includegraphics[width=\linewidth]{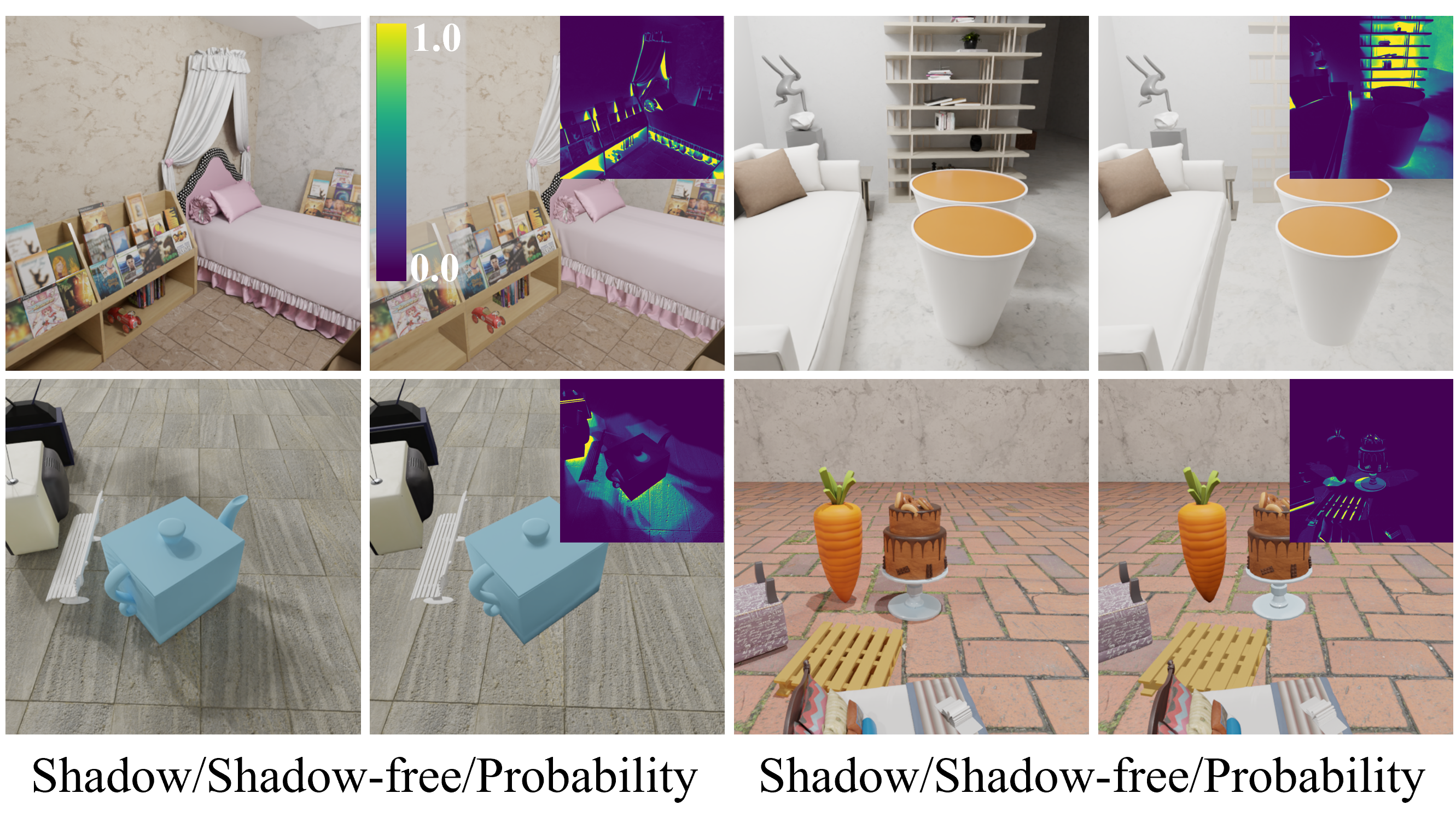}
\vspace{-0.5cm}
\caption{\textbf{Examples of our proposed dataset}. The rendered shadow and shadow-free image pairs and the generated shadow mask.}
\label{fig:dataset} 
\vspace{-0.2cm}
\end{figure}

The proposed INS dataset comprises realistic rendering results from various scenes featuring diverse object types, appearances, and intricate direct and indirect shadows. As illustrated in Fig.~\ref{fig:dataset}, the dataset is composed of two parts. The first part includes renderings from 3DFront scenes, where the rendering assets are sourced from the 3DFront dataset~\cite{fu20213dfront,fu20213dfuture}, which comprises 6813 synthetic indoor scenes. Drawing inspiration from \cite{paschalidou2021atiss}, we filter out nearly empty or overly crowded scenes, resulting in 6225 scenes. To further enhance the diversity of our dataset, we generate additional ``object composition scenes'' by randomly arranging objects from the ABO~\cite{collins2022abo} and Objaverse~\cite{deitke2023objaverse} datasets within an empty environment. In each scene, we place 3 to 5 objects within a fixed cubic space, ensuring no intersections, and illuminate them using both point and area lights. The cube is textured with six different materials from ambientCG~\cite{ambientCG}, as illustrated in the second row of Fig.~\ref{fig:dataset}. Our training data includes 20,000 image pairs from the 3DFront scenes and 10,000 image pairs from the ``object composition scenes''. While these ``object composition scenes'' do not match the realism of 3DFront scenes, they are effective in enhancing the quality of shadow removal in complex indoor environments, as demonstrated in the ablation study. See supp. for details on data preparation and scene generation.

% To further enhance the diversity of our dataset, we also add additional ``object composition scenes'' by randomly arranging objects from the ABO~\cite{collins2022abo} and Objaverse~\cite{deitke2023objaverse} datasets within an empty environment. We places 3 to 5 objects within a fixed cubic environment, without any intersection and using both point and area lights. The cubic applying six different textures from ambientCG~\cite{ambientCG}, as shown in the second row of Fig.~\ref{fig:dataset}. Although these additional scenes do not achieve the same level of realism in scene layout as those from the 3DFront dataset, they effectively improve the quality of removing complex shadows in real indoor environments. Please refer to the supp. for details on data preparation and scene generation.

% \begin{figure}[t]
% \centering
% \includegraphics[width=\linewidth]{./fig/scene_rearrangement.png}
% % \vspace{-0.5cm}
% \caption{\textbf{Examples of before/after rearrangement}. Our rearrangement procedure resolves unwanted collisions, as indicated by the red arrow.}
% \label{fig:scene_rearrangement} 
% % \vspace{-0.2cm}
% \end{figure}

During the rendering process, the efficient selection of materials, camera angles, and light sources is crucial. For rendering with the 3DFront dataset, we employ the default materials provided by 3DFront, selecting associated textures randomly from a predefined set. Camera poses are determined based on observed objects in the viewport and the shadow probability map. Once the camera poses are obtained, we select suitable light sources. Given the complex geometries of light sources in the 3DFront dataset, we focus on activating them based on the camera poses. To enhance our dataset, we also introduce extra point lights. The procedure for rendering ``object composition scenes'' is similar. Refer to the supp. for further information.

% ~\cite{wang2018stacked,le2019shadow,qu2017deshadownet} or shadow masks~\cite{vicente2016large,sun2023adaptive}

\textbf{Remark:} In comparison to existing shadow removal datasets (e.g., ISTD~\cite{wang2018stacked}, ISTD+~\cite{le2019shadow}, SRD~\cite{qu2017deshadownet}, and WRSD+~\cite{vasluianu2023wsrd}), our INS dataset presents four distinct advantages: 1. It encompasses a substantially larger number of shadow and shadow-free image pairs; 2. It integrates intricate and subtle indirect shadows into the dataset; 3. In our dataset, objects casting shadows can exist within the viewport, allowing for improved modelling of contextual information; 4. It provides precise shadow-free images, yielding ideal and noise-free shadow probability maps that facilitate the training of shadow detection.

The WRSD+~\cite{vasluianu2023wsrd} dataset, which uses both spotlight and diffuse flash lighting, also includes indirect shadows. However, these shadows are much simpler than those in our dataset, as they result from only few objects and a simple supporting plane. Additionally, since the shadow-free images in WRSD+ are captured with diffuse lighting, they are not entirely shadow-free and still contain some indirect shadows.

\subsection{Shadow Removal Network}

\begin{figure*}[t!]
\centering
\includegraphics[width=1.0\linewidth]{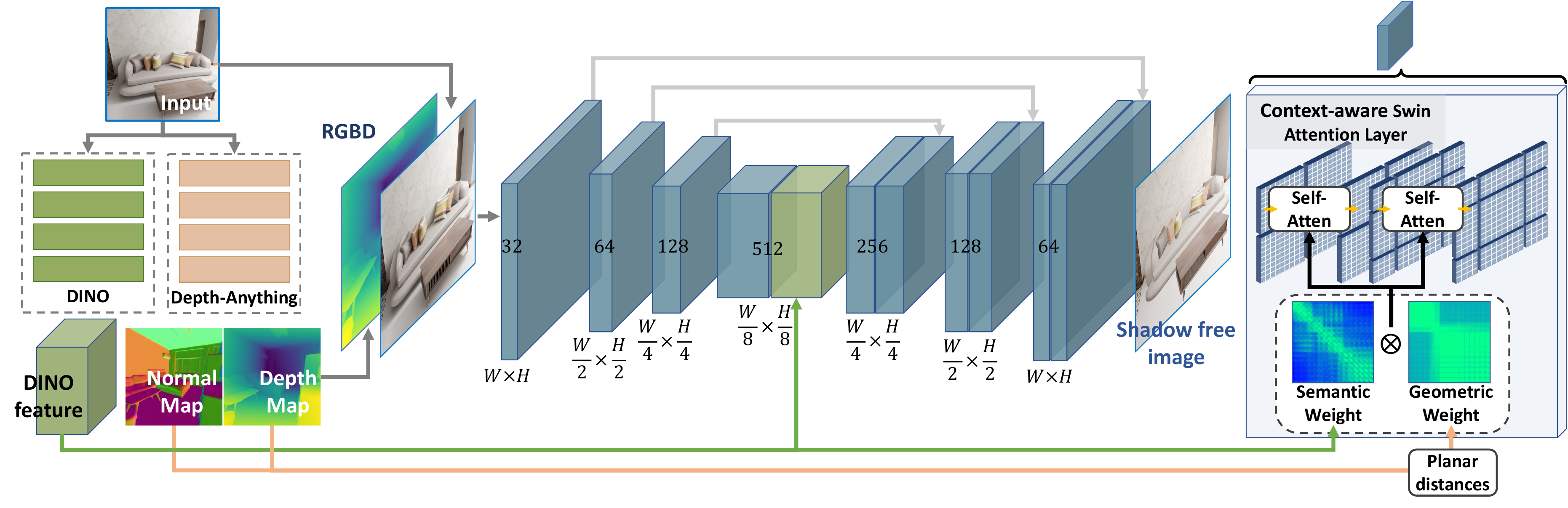}
\vspace{-0.7cm}
\caption{\textbf{Our proposed network.} For each input RGB image, we first extract its DINO features, depth, and normal map. Then, the RGB and depth images are inputted into the shadow removal network. The network includes several Context-aware Swin Attention (CSA) layers, each comprising two Swin self-attention blocks. Unlike traditional self-attention, our block explicitly involves semantic-aware and geometry-aware attention weights.}
\label{fig:flowchart} 
\vspace{-0.2cm}
\end{figure*}

We propose a shadow removal network, as illustrated in Fig.~\ref{fig:flowchart}. For each input image  $\mathbf{I}_s$, the network will output an estimated shadow-free image $\hat{\mathbf{I}}_f$. Our key observation is that shadows are inherently tied to geometry and semantics, resulting in darker shading than the surrounding non-shadow regions with similar semantics and materials. Hence, we propose the integration of geometric and semantic attention modules into the network. These modules are based on the pre-trained Depth-Anything-V2 network~\cite{yang2024depth} and DINO V2 network~\cite{oquab2023dinov2}. For each input image $\mathbf{I}_s$, our process first extracts its corresponding depth map $\mathbf{D}$ using Depth-Anything-V2 network and calculates its corresponding normal map $\mathbf{N}$, then retrieving its DINO feature map $\mathbf{F}$. The DINO features are trained on a large dataset in a self-supervised manner, capturing both materialistic and semantic implications \cite{sharma2023materialistic}.

\paragraph{Network Architecture.}

As depicted in Fig.~\ref{fig:flowchart}, the main component of our approach is a U-Net architecture~\cite{ronneberger2015u} consisting of a series of Swin Attention Blocks \cite{liu2021swin,wang2022uformer}, designed to extract and aggregate local and global context information. Specifically, the network comprises seven Context-aware Swin Attention (CSA) Layers, including a window attention and a shifted window attention. We concatenate the extracted DINO features in the bottleneck layer (the fifth layer). This addition of DINO features proves beneficial in mitigating ambiguity in identifying and removing shadows in complex indoor scenes.

In each CSA layer, the window attention is a window-based multi-head self-attention with relative position bias. Given the 2D feature maps $\mathbf{X}\in \mathbb{R} ^{C\times HW}$, we split $\mathbf{X}$ into non-overlapping windows $\mathbf{X}=\{\mathbf{X}^1,...,\mathbf{X}^{HW/{M^2}}\}$ of size $M\times M$, with a shifted size of $M/2 \times M/2$. To reduce the memory footprint, similar to ShadowFormer~\cite{guo2023shadowformer}, we substitute window attention with channel attention (CA) for the first two layers and the last two layers.

To enhance contextual perception in each CSA layer, we introduce semantic attention $\mathbf{W}^s$ and geometric attention $\mathbf{W}^g$ into the self-attention:
\begin{align}
\mathrm{Atten}(\mathbf{Q},\mathbf{K},\mathbf{V})&=\mathrm{softmax} \left( \frac{\mathbf{QK}^{\top}}{\sqrt{d_k}}\mathbf{W}+\mathbf{B} \right) \mathbf{V},
\\
\mathbf{W}&=\mathbf{W}^s\mathbf{W}^g,  \notag
\end{align}
where $\mathbf{Q}$, $\mathbf{K}$, $\mathbf{V}$ represent queries, keys, and values; $\mathbf{B}$ is the learnable relative position bias.

\paragraph{Semantic and Geometric Attention.}

For each non-overlapping windows $\mathbf{X}^i\in \mathbb{R} ^{C\times M^2}$, we calculate the attention weights ${\mathbf{W}^s, \mathbf{W}^g} \in \mathbb{R} ^{M^2 \times M^2}$, for every two pixels inside the window. The semantic attention weights are calculated based on the correlation of DINO features:
\begin{align}
\mathbf{W}^s&=\left[ \mathrm{dot}\left( \mathbf{F}_i,\mathbf{F}_j \right) \right] _{i,j\in \left[ 1,M^2 \right] \times \left[ 1,M^2 \right]},
\end{align}
where $i$ and $j$ represent two pixels in the window, the features are obtained using bilinear interpolation. 

% As illustrated in Fig. \ref{fig:weights}, the correlation among DINO features can be used to represent local semantic similarity.

The geometric attention is calculated based on the planar distance between two pixels:
\begin{align}
\mathbf{W}^g&=\left[ \mathrm{pdist}\left( \mathbf{D}_i,\mathbf{N}_i,\mathbf{D}_j,\mathbf{N}_j \right) \right] _{i,j},
\\
\mathrm{pdist}\left(\cdot\right)& =\frac{1}{2}\left| \mathbf{N}_i^{\top}\left( \pi \left( \mathbf{D}_i \right) -\pi \left( \mathbf{D}_j \right) \right) \right| \notag
\\ 
&+\frac{1}{2}\left| \mathbf{N}_j^{\top}\left( \pi \left( \mathbf{D}_i \right) -\pi \left( \mathbf{D}_j \right) \right) \right|, \notag
\end{align}
where $\mathbf{D}_i$ and $\mathbf{N}_i$ represent the depth and normal at pixel $i$, and $\pi(\cdot)$ is the back-projection function that transforms the depth into 3D points based on the pre-defined camera intrinsics. We opt for planar distances rather than Euclidean distances, as the latter tends to restrict feature aggregation to nearby regions. In contrast, planar distances incorporate the normal direction, considering points along planes or near-planar surfaces with short distances. This metric proves more stable in distinguishing foreground from background.

% have shaort distances along plane or near plane surfaces, which proves more effective in discerning foreground from background.

% \begin{figure}[t!]
% \centering
% \includegraphics[width=0.95\linewidth]{./fig/weights.pdf}
% \vspace{-0.1cm}
% \caption{Visualization of materialistic and geometric attentions for two key points within non-shadow region and shadow regions.}
% \vspace{-0.2cm}
% \label{fig:weights} 
% \end{figure}

\paragraph{Loss.}

During training, we employ Charbonnier loss~\cite{zamir2020learning} for shadow removal supervision:
\begin{align}
\mathcal{L}&=\sqrt{\left\| \mathbf{I}_f-\hat{\mathbf{I}}_f \right\| ^2+\epsilon ^2},
\end{align}
where $\mathbf{I}_f$ represents the ground-truth shadow-free image, $\hat{\mathbf{I}}_f$ is the estimated shadow-free image, and $\epsilon=10^{-3}$ is a constant in all the experiments. 

\paragraph{Training Details.} 

Our model is trained on a GPU server with four GeForce RTX 4090 GPUs using PyTorch 2.0.1~\cite{paszke2017automatic} with CUDA 11.7. We employ the Adam optimizer~\cite{Kingma2015AdamAM} for training. The initial learning rate is set to $2\times10^{-4}$ and adjusted using a cosine annealing scheduler~\cite{loshchilov2016sgdr}. Additional details can be found in the supplementary.

% Our model is trained on a GPU server with four GeForce RTX 4090 GPUs using PyTorch 1.9.0~\cite{paszke2017automatic} with CUDA 11.1. The batch size was consistently set at 8, and each batch consisted of $512\times512$ patches for our INS dataset and $480\times480$ cropped patches for all other datasets. The window size for the Swin attention layer was set as 16 and 15, respectively. In our experiments, we concluded the training procedure after 40 epochs for the INS dataset and 500 epochs for other datasets. Please refer to the supplementary material for more details.

% We employ the distributed data parallel (DDP) to accelarate the training process. For optimization, we utilize the Adam optimizer~\cite{Kingma2015AdamAM}. The learning rate is initially set to $1\times10^{-3}$ and is adjusted using a cosine annealing~\cite{loshchilov2016sgdr} scheduler. Additionally, during training, we incorporated simple data augmentation methods, including random rotation (ranging from $-20$ to $+20$ degrees) and horizontal flipping. We believe that more sophisticated data augmentation methods can also be explored and integrated for further improvement.

\section{Experiments}

\begin{table*}[t]
\small
\center
\begin{tabular}{l c c c c c c c c c}
\hline
\multirow{2}{*}{Method} & \multirow{2}{*}{Year} & \multicolumn{2}{c}{ISTD Dataset} & \multicolumn{2}{c}{ISTD+ Dataset} & \multicolumn{2}{c}{SRD Dataset} & \multicolumn{2}{c}{WSRD+ Dataset} \\ \cmidrule(lr){3-4}\cmidrule(lr){5-6}\cmidrule(lr){7-8}\cmidrule(lr){9-10}
& & {PSNR$\uparrow$} & {SSIM$\uparrow$} & {PSNR$\uparrow$} & {SSIM$\uparrow$} & {PSNR$\uparrow$} & {SSIM$\uparrow$} & {PSNR$\uparrow$} & {SSIM$\uparrow$} \\  
\hline
DSC~\cite{hu2019direction} & 2019 & 29.00 & 0.944 & 25.66 & 0.956 & 29.05 & 0.940 & --- & --- \\ 
DHAN~\cite{cun2020towards} & 2020 & 29.11 & 0.954 & 25.66 & 0.956 & 30.74 & 0.958 & 22.39 & 0.796\\ 
Fu et al.~\cite{fu2021auto} & 2021 & 26.30 & 0.835 & 28.40 & 0.846 & 28.52 & 0.932 & 21.66 & 0.752\\
BMNet~\cite{zhu2022bijective} & 2022 & 28.53 & 0.952 & \cellcolor{colorSnd}{32.22} &  \cellcolor{colorSnd}{0.965} & 28.34 & 0.943 & 24.75 & 0.816\\ 
TBRNet~\cite{liu2023shadow} & 2023 & 28.77 & 0.928 &  \cellcolor{colorTrd}{31.91} &  \cellcolor{colorTrd}{0.964} & 31.83 & 0.953 & --- & --- \\ 
ShadowFormer~\cite{guo2023shadowformer} & 2023 & \cellcolor{colorTrd}{29.90} & \cellcolor{colorSnd}{0.960} & 31.39 & 0.946 & 30.58 & 0.958 & \cellcolor{colorTrd}{25.44} & \cellcolor{colorTrd}{0.820}\\ 
DMTN~\cite{liu2023decoupled} & 2023 & 29.05 & \cellcolor{colorTrd}{0.956} & 31.72 & 0.963 & \cellcolor{colorSnd}{32.45} & \cellcolor{colorTrd}{0.964} & --- & --- \\
ShadowDiffusion~\cite{guo2023shadowdiffusion} & 2023 & \cellcolor{colorSnd}{30.09} & 0.918 & 31.08 & 0.950 & \cellcolor{colorTrd}{31.91} & \cellcolor{colorSnd}{0.968} & --- & --- \\ 
ShadowRefiner~\cite{dong2024shadowrefiner} & 2024 & --- & --- & --- & --- & --- & --- & \cellcolor{colorSnd}{26.04} & \cellcolor{colorSnd}{0.827}\\ 
Ours & --- & \cellcolor{colorFst}{30.45} & \cellcolor{colorFst}{0.964} & \cellcolor{colorFst}{33.34} & \cellcolor{colorFst}{0.970} & \cellcolor{colorFst}{32.87} & \cellcolor{colorFst}{0.969} & \cellcolor{colorFst}{26.07} & \cellcolor{colorFst}{0.835}  \\
\hline
\hline
Fu et al.~\cite{fu2021auto} + GM &        2021 & 27.19 & 0.945 & 29.45 & 0.861 & 29.24 & 0.938 & --- & --- \\ 
Zhu et al.~\cite{zhu2022efficient} + GM & 2022 & 29.85 & 0.960 & --- & --- & 32.05 & 0.965 & --- & --- \\
BMNet~\cite{zhu2022bijective} + GM &      2022 & 30.28 & 0.959 & 33.98 & \cellcolor{colorSnd}{0.972} & 31.97 & 0.965 & --- & --- \\ 
ShadowFormer~\cite{guo2023shadowformer} + GM & 2023 & \cellcolor{colorSnd}{32.21} & \cellcolor{colorSnd}{0.968} & \cellcolor{colorSnd}{35.46} & \cellcolor{colorTrd}{0.971} & {32.90} & 0.958 & --- & --- \\ 
DMTN~\cite{liu2023decoupled} + GM & 2023 & 30.42 & \cellcolor{colorTrd}{0.965} & 33.68 & \cellcolor{colorTrd}{0.971} & \cellcolor{colorTrd}{33.77} & \cellcolor{colorTrd}{0.968} & --- & --- \\
ShadowDiffusion~\cite{guo2023shadowdiffusion} + GM & 2023 & \cellcolor{colorFst}{32.33} & \cellcolor{colorFst}{0.969} & \cellcolor{colorFst}{35.72} & 0.969 & \cellcolor{colorFst}{34.73} & \cellcolor{colorSnd}{0.970} & --- & --- \\
Ours + GM & --- & \cellcolor{colorTrd}{31.56} & \cellcolor{colorTrd}{0.965} & \cellcolor{colorTrd}{34.20} & \cellcolor{colorFst}{0.973} & \cellcolor{colorSnd}{34.56} & \cellcolor{colorFst}{0.977} & --- & --- \\
\hline
\end{tabular}
\caption{\textbf{Quantitative comparisons on ISTD, ISTD+, SRD, and WSRD+ datasets.} Best results are highlighted as \colorbox{colorFst}{1st}, \colorbox{colorSnd}{2nd} and \colorbox{colorTrd}{3rd}. +GM: using ground-truth shadow masks.}
\vspace{-0.2cm}
\label{tab:comparison_ISTD}
\end{table*}

% 31.56 0.965
% 34.20 0.973
% 34.56 0.977
 
\begin{table}[t]
\setlength\tabcolsep{4pt}
\small
\center
\begin{tabular}{l c c c}
\hline
\multicolumn{3}{c}{INS Dataset}\\
\hline
\multirow{2}{*}{Method} & INS testing & Real testing \\
\cmidrule(lr){2-2}\cmidrule(lr){3-3} & {PSNR$\uparrow$}/{SSIM$\uparrow$} & {PSNR$\uparrow$}/{SSIM$\uparrow$} \\  
\hline
DHAN~\shortcite{cun2020towards} & 27.84/0.963 & 35.05/0.993 & \\ 
Fu et al.~\shortcite{fu2021auto} & 27.91/0.957 & 36.64/0.994 & \\ 
BMNet~\shortcite{zhu2022bijective} & 27.90/0.958 & 36.65/0.994 & \\ 
ShadowFormer~\shortcite{guo2023shadowformer} & 28.62/0.963 & \colorbox{colorTrd}{36.99/0.994} & \\ 
DMTN~\shortcite{liu2023decoupled} & \colorbox{colorTrd}{28.83/0.969} & 35.83/0.993 & \\
ShadowDiffusion~\shortcite{guo2023shadowdiffusion} & \colorbox{colorSnd}{29.12/0.966} & \colorbox{colorSnd}{36.91/0.994}
 &\\
Ours & \colorbox{colorFst}{30.38/0.973} & \colorbox{colorFst}{38.34/0.995} & \\ 
\hline
\end{tabular}
\caption{\textbf{Quantitative comparisons on our INS dataset and real captured images.} Best results are highlighted as \colorbox{colorFst}{1st}, \colorbox{colorSnd}{2nd} and \colorbox{colorTrd}{3rd}.}
\label{tab:comparison_INS}
\end{table}

\paragraph{Datasets and Evaluation Metrics.} 

\begin{figure}[t!]
\centering
\includegraphics[width=1.0\linewidth]{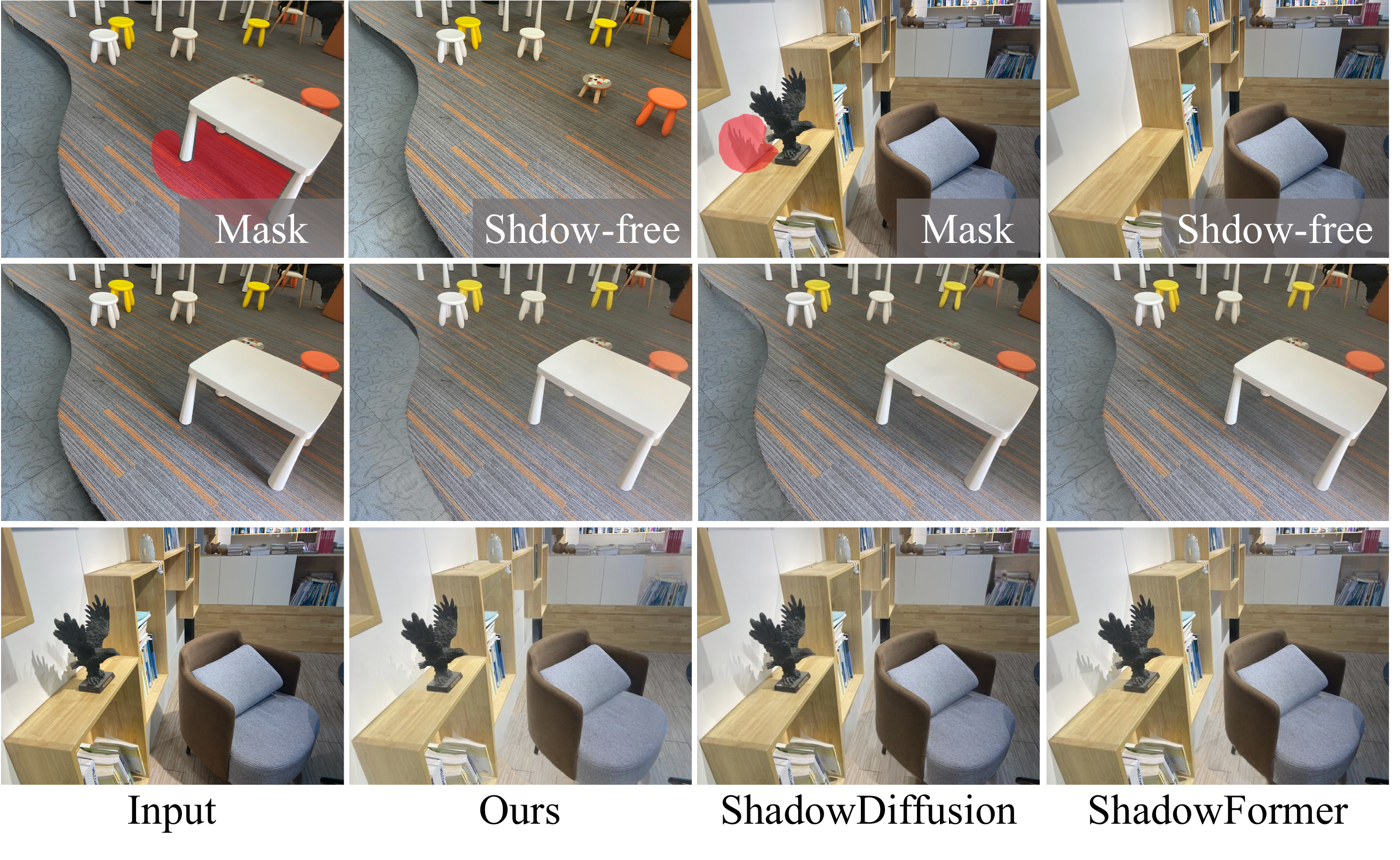}
\vspace{-0.4cm}
\caption{\textbf{Real data comparisons.} For the real captured testing data, our method excels in thoroughly removing complex indirect shadows. The annotated masks are highlighted in red, and the corresponding quantitative comparisons are presented in Table \ref{tab:comparison_INS}.}
\vspace{-0.2cm}
\label{fig:ins_real_testing} 
\end{figure}

We conducted our experiments on ISTD~\cite{wang2018stacked}, ISTD+~\cite{le2019shadow}, SRD~\cite{qu2017deshadownet}, WRSD+~\cite{vasluianu2023wsrd}), and the proposed INS dataset. We also evaluated the generalizability of our method in real indoor settings. As illustrated in Fig. \ref{fig:ins_real_testing}, we captured pairs of images (shadow and shadow-free) and manually annotated some shadow regions, which can be shadow-free in the counterpart image.

% We conducted our experiments on ISTD~\cite{wang2018stacked}, ISTD+~\cite{le2019shadow}, SRD~\cite{qu2017deshadownet}, WRSD+~\cite{vasluianu2023wsrd}), and the proposed INS dataset. The evaluations for these datasets were carried out using models trained on their respective training sets without the inclusion of any additional training data. As shown in Fig. \ref{fig:comparison}, the first three datasets primarily consist of real captured shadowed and shadow-free images in outdoor scenes and often involve direct shadows. WRSD+~\cite{vasluianu2023wsrd} contains shadows of objects on a plane, illuminated by a spotlight and diffuse lighting. In contrast, our INS dataset contains images with indoor indirect shadows. We also evaluated the generalizability of our method in real indoor settings. As illustrated in Fig. \ref{fig:ins_real_testing}, we captured pairs of images (shadow and shadow-free) and manually annotated some shadow regions, which can be shadow-free in the counterpart image.

We evaluated images with a resolution of 256 × 256, following previous methods \cite{fu2021auto,le2020shadow,guo2023shadowformer}. We report results using the Peak Signal-to-Noise Ratio (PSNR) and the Structure Similarity Index Measure (SSIM)~\cite{wang2004image}, adopting the MATLAB evaluation codes as provided by Zhu et al. \cite{zhu2022efficient}. For the WRSD+ \cite{vasluianu2023wsrd} dataset, since it does not provide testing data, we used its evaluation data and the evaluation code provided by the NTIRE 2024 Image Shadow Removal Challenge~\cite{vasluianu2024ntire} for comparison.

\begin{figure*}[t!]
\centering
\includegraphics[width=0.95\linewidth]{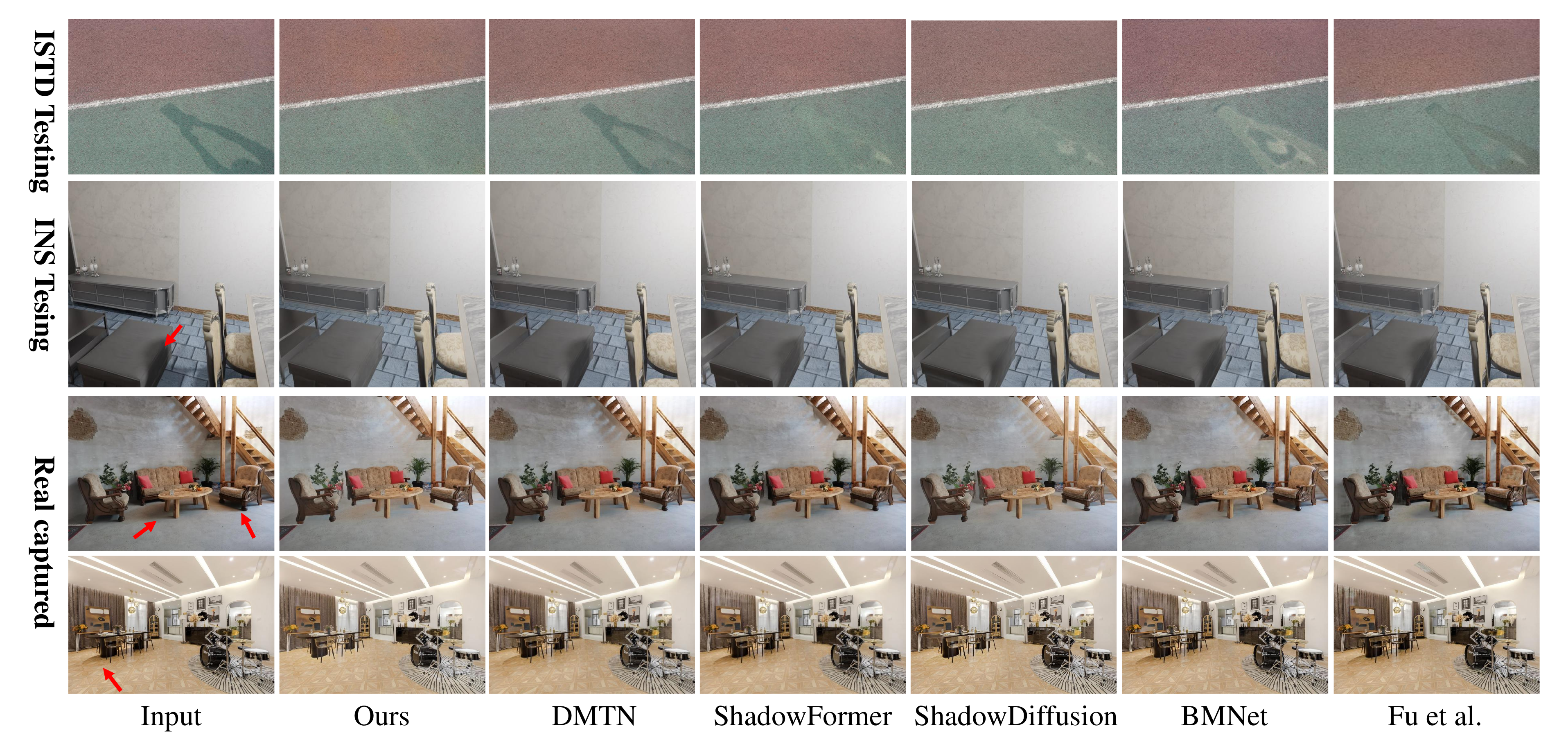}
%\vspace{-0.5cm}
\caption{\textbf{Comparisons with SOTA shadow removal methods show improved quality of our method.} Comparisons with DMTN \cite{liu2023decoupled}, ShadowFormer \cite{guo2023shadowformer},  ShadowDiffusion \cite{guo2023shadowdiffusion}, BMNet \cite{zhu2022bijective} and Fu et al. \cite{fu2021auto} on both outdoor and indoor scenes. Our method demonstrates more comprehensive shadow removal, even in complex scenes.}
\vspace{-0.2cm}
\label{fig:comparison} 
\end{figure*}

\subsection{Comparisons}

We compare our method with nine state-of-the-arts shadow removal methods, including the DSC~\cite{hu2019direction}, DHAN~\cite{cun2020towards}, Fu et al.~\cite{fu2021auto}, Zhu et al.~\cite{zhu2022efficient}, BMNet~\cite{zhu2022bijective}, ShadowFormer\cite{guo2023shadowformer}, DMTN~\cite{liu2023decoupled}, ShadowDiffusion~\cite{guo2023shadowdiffusion}, and ShadowRefiner~\cite{dong2024shadowrefiner} both quantitatively (Table \ref{tab:comparison_ISTD} and \ref{tab:comparison_INS}) and qualitatively (Fig. \ref{fig:comparison} and \ref{fig:ins_real_testing}). All comparisons use the results reported in the original papers or the original authors' implementations and hyperparameters.

Note that Fu et al.~\cite{fu2021auto}, Zhu et al.~\cite{zhu2022efficient}, BMNet~\cite{zhu2022bijective}, ShadowFormer~\cite{guo2023shadowformer}, DMTN~\cite{liu2023decoupled}, and ShadowDiffusion~\cite{guo2023shadowdiffusion} make use of explicit shadow masks as input, treating them as auxiliary information. Typically, these methods present their results using ground-truth shadow masks provided by the dataset as the default input, which is not available in real-life scenarios. In our approach, alongside reporting the shadow removal performance using ground-truth masks, denoted as ``+ GM'', we also evaluate their performance using imperfectly detected shadow masks from~\cite{zhu2021mitigating}.

%, similar to BMNet~\cite{zhu2022bijective}. 

As shown in Table~\ref{tab:comparison_ISTD}, our method achieves the highest PSNR and SSIM scores on ISTD, ISTD+, SRD, and INS datasets without ground-truth shadow masks. Even when compared with other methods utilizing ground-truth shadow masks, our approach, which does not rely on such masks, achieves the third-best results on the ISTD dataset, surpassed only by ``ShadowDiffusion~\cite{guo2023shadowdiffusion} + GM'' and ``ShadowFormer \cite{guo2023shadowformer} + GM''. When concatenated with ground-truth shadow masks as input, our method also demonstrates a significant improvement in PSNR performance on the ISTD, ISTD+, and SRD datasets. 

As demonstrated in Fig. \ref{fig:comparison}, our method excels in removing direct shadows in the ISTD dataset and both direct and indirect shadows in synthetic testing data (INS testing) and real captured images. For the WRSD+ dataset, our method also outperforms others, including ShadowRefiner~\cite{dong2024shadowrefiner}. The relatively low PSNR scores for all methods on the WRSD+ dataset can be attributed to exposure differences between the input and ground-truth images. As shown in Fig~\ref{fig:comparison}, although our synthetic dataset significantly enhances the performance of DMTN, ShadowFormer, and ShadowDiffusion in reducing complex shadows in indoor scenes, they still struggle to eliminate shadows in these areas completely. This limitation may be attributed to these methods lacking explicit and high-quality semantic and geometric perception. We provide additional results in the supplementary material.

To evaluate the generalizability of our method, we performed additional assessments using a collection of 100 captured pairs of shadow and ``shadow-free'' images, each at a resolution of $640\times480$. Notably, our network is trained on the synthetic INS dataset. The quantitative comparisons are detailed in the third column of Table \ref{tab:comparison_INS}, where our method significantly outperforms alternative approaches. As depicted in Fig.~\ref{fig:ins_real_testing}, our method outperforms competing methods in scenarios involving shadows with multiple light sources and indirect illumination.

\paragraph{Runtime comparison.} In Table~\ref{tab:runtime}, we present a comparison of inference time (for a $640 \times 480$ image). Due to the involvement of the Depth-Anything-V2 network~\cite{yang2024depth} and the DINO V2 network~\cite{oquab2023dinov2}, our method exhibits higher computational complexity compared to lightweight methods like ShadowFormer~\cite{guo2023shadowformer}. However, our method is faster than diffusion-based methods such as ShadowDiffusion~\cite{guo2023shadowdiffusion}.

\begin{table}[t]
\setlength\tabcolsep{4pt}
\small
\center
\begin{tabular}{c c c c c}
\hline
\multirow{1}{*}{} & Ours & DMTN & ShadowFormer & ShadowDiffusion \\
\hline
Time & 120.1 & 82.6 & 43.7 & 506.9 \\
\hline
\end{tabular}
\caption{\textbf{Runtime comparisons (in milliseconds).}}
\label{tab:runtime}
\end{table}

% Below, we provide the timing requirements (640x480 image). We will include a timing table and a related discussion in the experiments section.
%         Ours (network w/ DINO+Depth est.)   DMTN    ShadowFormer   ShadowDiffusion
% time    74.3+45.8ms                         82.6    43.7           506.9
% #para   24.5+80.0+24.2M                     45.6    11.4           55.49

\begin{table}[t]
\setlength\tabcolsep{4pt}
\small
\center
\begin{tabular}{c c c c}
\hline
\multirow{2}{*}{Dataset} & INS Testing & Real captured & WRSD+ \\
& PSNR$$/SSIM$$ & PSNR$$/SSIM$$ & PSNR$$/SSIM$$ \\
\hline
Full & \textbf{30.38/0.973} & \textbf{38.34/0.995} & \textbf{26.07/0.835} \\
W/o depth & 29.95/0.973 & 38.03/0.995 & 25.73/0.828 \\ 
W/o DINO & 29.31/0.966 & 37.06/0.994 & 23.31/0.797 \\
W/o $\mathbf{W}^s$ & 30.01/0.972 & 38.18/0.995 & 25.96/0.834 \\ 
W/o $\mathbf{W}^g$ & 30.22/0.972 & 37.82/0.995 & 25.77/0.832 \\
W/o obj scenes & 29.32/0.967 & 37.75/0.995 & --- \\
\hline
\end{tabular}
\caption{\textbf{Ablation studies.} W/o depth: only RGB input. W/o DINO: without DINO concatenation.}
\vspace{-0.5cm}
\label{tab:ablation}
\end{table}

\subsection{Ablation Study}

To validate our model designs, we conducted ablation studies on the proposed semantic and geometric attention weights, depth concatenation, and DINO feature concatenation. The ``INS testing'' and ``real captured'' are trained on our synthetic training data. The ``WRSD+'' are trained and evaluated using the WRSD+ dataset. The performance is reported in Table \ref{tab:ablation}. The semantic and geometric attention mechanisms enhance PSNR/SSIM, and the DINO concatenation is a critical component as PSNR/SSIM drop severely when it is removed. We think this is due to its ability to provide semantic and materialistic information, which are crucial for effectively identifying shadows and mitigating ambiguities between shadow and non-shadow areas with similar appearances. Furthermore, as detailed in Table \ref{tab:ablation}, the concatenation of monocular depth is also crucial for our shadow removal approach. Including monocular depth provides meaningful geometric cues for precise shadow removal. We also evaluate the impact of our additional ``object composition scenes''. As indicated in the last row of Table~\ref{tab:ablation}, the PSNR drops significantly without ``object composition scenes'' (w/o obj scenes). Please refer to the supp. for qualitative results related to the ablation study.

\section{Conclusion}

This paper introduces a novel neural approach for effectively eliminating both direct shadows and subtle, soft indirect shadows. Extensive experiments show that our approach surpasses state-of-the-art shadow removal methods. 

% Our network architecture is built upon the U-Net framework with Swin Attention Blocks. We enhance its contextual perception by introducing explicit materialistic and geometric attention mechanisms. Moreover, we created a synthetic dataset using path-tracing techniques with intricate indoor scenes featuring various shadows. We also captured diverse testing images with annotated local masks to evaluate our method better. 

\noindent{\textbf{Limitation:}} Our method may fail to eliminate vast shadow regions, such as those covering entire surfaces or objects. 

% Furthermore, given that 3D-Front primarily consists of furniture, our synthetic dataset could benefit from increased diversity in smaller objects.

% to detect shadows from raw images. Our network has
% a novel adaptive illumination mapping module to predict
% sRGB images of different intensity ranges, and a shadow
% detection module to exploit such illumination information
% to detect shadows. We have also proposed a novel feedback
% mechanism to guide the illumination mapping process in
% a shadow-aware manner. To facilitate the learning process,
% we have constructed a new dataset with raw images and corresponding shadow masks. Extensive experiments demonstrate that our method outperforms state-of-the-art shadow
% detection approaches.

%\newpage

\section{Acknowledgments}

We thank the anonymous reviewers for their professional and constructive comments. Jiamin Xu is partially supported by the National Key R\&D Program of China under Grant No.~2023YFB3309100, NSFC grant No.~62302134, Zhejiang Provincial Natural Science Foundation under Grant No.~LQ24F020031, the Fundamental Research Funds for the Provincial Universities of Zhejiang under Grant No.~GK249909299001-021. Renshu Gu is partially supported by NSFC grant No.~62202130 and U22A2033. Weiwei Xu is partially supported by NSFC grant No.~61732016.

% We thank the anonymous reviewers for their constructive comments. Weiwei Xu is partially supported by NSFC grant No.~61732016. Jiamin Xu is partially supported by NSFC grant No.~62302134, Zhejiang Provincial Natural Science Foundation under Grant No.~LQ24F020031. 

% This research was supported by the National Natural Science Foundation of China (No. 62302134), the Zhejiang Provincial Natural Science Foundation of China (No. LQ24F020031), the Fundamental Research Funds for the Provincial Universities of Zhejiang (No. GK249909299001-021).
% 

% This research was supported by NSFC grant No.~62302134, Zhejiang Provincial Natural Science Foundation of China under Grant No.~LQ24F020031
% Uncomment the following to link to your code, datasets, an extended version or similar.
%
% \begin{links}
%     \link{Code}{https://blackjoke76.github.io/Projects/OmniSR/}
%     % \link{Datasets}{https://blackjoke76.github.io/Projects/OmniSR/}
%     % \link{Extended version}{https://aaai.org/example/extended-version}
% \end{links}

\bibliography{main}

% \newpage
% \clearpage
% \input{sec/checklist}

\end{document}